\documentclass[letterpaper]{article}
\usepackage{ijcai19}
\usepackage{colonequals}
\usepackage{cmap}
\usepackage{amsmath,amsfonts,amssymb}
\usepackage{graphicx}
\usepackage{microtype}
\usepackage{xspace}
\usepackage[ruled,vlined]{algorithm2e}
\usepackage{wrapfig}
\usepackage{subfigure}
\usepackage{xcolor}
\usepackage{booktabs,colortbl,tabularx}
\usepackage{arydshln}
\usepackage{nicefrac}
\usepackage{times}
\usepackage{courier}
\usepackage{helvet}
\usepackage{paralist}
\usepackage{tikz}
\usepackage{multirow}
\usepackage{url}
\usepackage{pgfplots}
\usepackage[disable]{todonotes}
\usepackage{theorem,balance}
\usepackage{acronym}
\usepackage{bm}
\usepackage{mdframed}
\usepackage{subfigure}

\usetikzlibrary{patterns,arrows,arrows.meta,calc,shapes,shadows,decorations.pathmorphing,decorations.pathreplacing,automata,shapes.multipart,positioning,shapes.geometric,fit,circuits,trees,shapes.gates.logic.US,fit}
\usetikzlibrary{backgrounds,scopes}

\tikzstyle{decision} = [diamond, draw, fill=blue!20, 
text width=4.5em, text badly centered, node distance=3cm, inner sep=0pt]
\tikzstyle{block} = [rectangle, draw, fill=blue!20, 
text width=5em, text centered, rounded corners, minimum height=4em]
\tikzstyle{line} = [draw, -latex']
\tikzstyle{cloud} = [draw, ellipse,fill=red!20, node distance=3cm,
minimum height=2em]
\def\checkmark{\tikz\fill[scale=0.6](0,.35) -- (.25,0) -- (1,.7) -- (.25,.15) -- cycle;} 

\DontPrintSemicolon
\SetKwInOut{Input}{Input}\SetKwInOut{Output}{Output}

\SetCommentSty{mycommfont}

\tikzset{>=stealth}
\tikzset{state/.append style={inner sep=2pt,minimum size=2pt}}

\newtheorem{definition}{Definition}

\usepackage{array}
\newcolumntype{H}{>{\setbox0=\hbox\bgroup}c<{\egroup}@{}}

\newcommand{\ie}{i.\,e.,\@\xspace}

\newcommand{\eg}{e.\,g.\@\xspace}

\newcommand{\tool}[1]{\textrm{#1}\xspace}

\newcommand{\p}{\ensuremath{\mathbb{P}}}
\newcommand{\pr}{\ensuremath{\mathrm{Pr}}}

\newcommand{\reachPropSymbol}{\varphi}
\newcommand{\ltlformula}{\psi}

\newcommand{\rew}{\ensuremath{r}}

\newcommand{\finally}{\lozenge}
\newcommand{\R}{\mathbb{R}}

\newcommand{\Ireal}{[0,\, 1]\subseteq\mathbb{R}}  % real interval [0,1]
\newcommand{\Ex}{\ensuremath{\mathbb{E}}\xspace}        % Expectations
\newcommand{\Distr}{\mathit{Distr}}

\newcommand{\distDom}{X}

\newcommand{\distFunc}{\mu}
\newcommand{\distDomElem}{x}
\DeclareMathOperator{\supp}{supp}

\newcommand{\Always}{\Box\,}
\newcommand{\Finally}{\lozenge\,}
\newcommand{\Ever}{\Diamond\,}
\newcommand{\Next}{\bigcirc\,}
\newcommand{\Until}{\mbox{$\, {\sf U}\,$}}

\newcommand{\mdp}{M}
\newcommand{\MdpInit}[1][]{\ensuremath{\mdp{#1}=\allowbreak(S{#1},\Act,\probmdp{#1})}}

\newcommand{\probmdp}{\mathcal{P}}
\newcommand{\probdtmc}{P}

\newcommand{\Strategy}{\sched} % Strategy/policy of an MDP
\newcommand{\DTMCgnd}{M^\Strategy}

\newcommand{\fsc}{\ensuremath{\mathcal{A}}}

\newcommand{\ObsSym}{{Z}}
\newcommand{\ObsFun}{{O}}
\newcommand{\obs}{\ensuremath{z}}
\newcommand{\PomdpInit}[1][]{\pomdp{#1}=(\mdp{#1},\ObsSym{#1},\ObsFun{#1})}
\newcommand{\pomdp}{\mathcal{M}}

\newcommand{\states}{\ensuremath{S}}

\newcommand{\sched}{\ensuremath{\sigma}}
\newcommand{\Sched}{\ensuremath{{\Sigma}}}
\newcommand{\osched}{\ensuremath{\mathit{\sigma}}}
\newcommand{\oSched}{\ensuremath{\Sigma}}

\newcommand{\Act}{\ensuremath{\mathit{Act}}}
\newcommand{\act}{\ensuremath{a}}

\newcommand{\obsSeq}{\mathsf{ObsSeq}_{\mathit{fin}}}
\newcommand{\obsSeqFin}{\obsSeq}
\newcommand{\pathset}{\mathsf{Paths}}

\newcommand{\pathsfin}{\pathset_{\mathit{fin}}}

\newcommand{\last}[1]{\mathrm{last}(#1)}

\DeclareMathAlphabet{\mathpzc}{OT1}{pzc}{m}{it}
\def\presuper#1#2%
  {\mathop{}%
   \mathopen{\vphantom{#2}}^{#1}%
   \kern-\scriptspace%
   #2}

\newcommand{\gridScale}{1} %needs to be predefined here

\newcommand{\fillGridAt}[3]{
	\node [xshift=.5*\gridScale cm,yshift=.5*\gridScale cm] at (#1,#2){#3};	
}

\newcommand{\gridsizeparam}{\ensuremath{c}}

\newcommand{\RNNfun}{\ensuremath{\hat{\sched}}}

\begin{document}

\pdfinfo{
   /Title  (Counterexample-Guided Strategy Improvement for POMDPs Using Recurrent Neural Networks)
   /Author (Author names omitted for submission)
}

\title{Counterexample-Guided Strategy Improvement for POMDPs \\ Using Recurrent Neural Networks\thanks{This work was partially supported by the grants DARPA D19AP00004 and ONR N00014-18-1-2829.}}

%\author{Author names omitted for submission}
\author{Steven Carr\textsuperscript{1}, Nils Jansen\textsuperscript{2},
        Ralf Wimmer\textsuperscript{3,4}, \\
        \Large\textbf{Alexandru C. Serban,\textsuperscript{2}, Bernd Becker\textsuperscript{3} and Ufuk Topcu\textsuperscript{1}}\vspace{0.3cm} \\
  \textsuperscript{1}The University of Texas at Austin \\
  \textsuperscript{2}Radboud University, Nijmegen, The Netherlands \\
  \textsuperscript{3}Albert-Ludwigs-Universit\"at Freiburg, Freiburg im Breisgau, Germany \\
  \textsuperscript{4}Concept Engineering GmbH, Freiburg im Breisgau, Germany
}

\maketitle
\acrodef{LSTM}[LSTM]{long short-term memory}
\acrodef{POMDP}[POMDP]{partially observable Markov decision process}
\acrodefplural{POMDP}[POMDPs]{partially observable Markov decision processes}
\acrodef{RNN}[RNN]{recurrent neural network}
\acrodef{FSC}[FSC]{finite-state controller}
\acrodef{DTMC}[MC]{discrete-time Markov chain}
\acrodef{LP}[LP]{linear programming}
\acrodef{FSC}[FSC]{finite-state controller}
\acrodef{MDP}[MDP]{Markov decision process}
\acrodefplural{MDP}[MDPs]{Markov decision processes}
\acrodef{LTL}[LTL]{linear-time temporal logic}
\acrodef{NN}[NN]{neural network}

\begin{abstract}
	We study strategy synthesis for \acp{POMDP}.
	The particular problem is to determine strategies that provably adhere to (probabilistic) temporal logic constraints.
	This problem is computationally intractable and theoretically hard.
	We propose a novel method that combines techniques from machine learning and formal verification.
	First, we train a \ac{RNN} to encode POMDP strategies.
	The \ac{RNN} accounts for memory-based decisions without the need to expand the full belief space of a \ac{POMDP}.
	Secondly, we restrict the \ac{RNN}-based strategy to represent a finite-memory strategy and implement it on a specific \ac{POMDP}.
	For the resulting finite Markov chain, efficient formal verification techniques provide provable guarantees against temporal logic specifications.
	If the specification is not satisfied, counterexamples supply diagnostic information.
	We use this information to improve the strategy by iteratively training the \ac{RNN}.
	Numerical experiments show that the proposed method elevates the state of the art in \ac{POMDP} solving by up to three orders of magnitude in terms of solving times and model sizes. 
\end{abstract}

\section{Introduction}
\label{sec:introduction}
\acresetall
Autonomous agents that make decisions under uncertainty and incomplete information can be mathematically represented as \acp{POMDP}.
In this setting, while an agent makes decisions within an environment, it obtains \textit{observations} and infers the likelihood of the system being in a certain state, known as the \textit{belief} state.
\acp{POMDP} are effective in modeling a number of real-world applications, see for instance~\cite{kaelbling1998planning}.
%such as motion planning with limited visibility~\cite{thrun2005probabilistic} and sampling terrain with noisy sensors~\cite{smith2004heuristic}.
Traditional \ac{POMDP} problems typically seek to compute a strategy that maximizes a cumulative reward over a finite horizon.

However, the agent's behavior is often required to obey more complicated specifications.
For example, reachability, liveness or, more generally, specifications expressed in temporal logic (\eg~LTL~\cite{Pnueli77}) describe tasks that cannot be expressed using reward functions~\cite{littman2017environment}.

Strategy synthesis for POMDPs is a difficult problem, both from the theoretical and the practical perspective.
For infinite- or indefinite-horizon problems, computing an optimal strategy is undecidable~\cite{MadaniHC99}.
Optimal action choices depend on the whole history of observations and actions, thus requiring an infinite amount of memory.
When restricting the specifications to maximize accumulated rewards over a finite horizon and limiting the available memory, computing an optimal strategy is PSPACE-complete~\cite{papadimitriou1987complexity}.
This problem is, practically, intractable even for small instances~\cite{meuleau1999learning}.
Moreover, even when strategies are restricted to be \emph{memoryless}, finding an optimal strategy within this set is still NP-hard~\cite{VlassisLB12}.
For more general specifications like LTL properties, synthesis of strategies with limited memory is even harder, namely EXPTIME-complete~\cite{chatterjee2015qualitative}).

The intractable nature of finding exact solutions in these problems gave rise to approximate ~\cite{hauskrecht2000value}, point-based~\cite{pineau2003point}, or Monte-Carlo-based~\cite{silver2010monte} methods.
However, none of these approaches provides guarantees for given temporal logic specifications.
The tool PRISM-POMDP \cite{NPZ17} does so by approximating the belief space into a fully observable belief MDP, but is restricted to small examples.

Although strategy synthesis for POMDPs is hard, an available \emph{candidate strategy} resolves the nondeterminism and partial observability for a POMDP and yields a so-called induced \ac{DTMC}.
For this simpler model, verification methods are capable to efficiently certify temporal logic constraints and reward specifications for billions of states~\cite{BK08}.
Tool support is available via probabilistic model checkers such as PRISM~\cite{KNP11} or Storm~\cite{DBLP:conf/cav/DehnertJK017}.

There remains a dichotomy between directly synthesizing an optimal strategy and the efficient verification of a candidate strategy.
The key questions are (1) how to generate a ``good'' strategy in the first place and (2) how to improve a strategy if verification refutes the specification.
Machine learning and formal verification techniques  address these questions separately.
In this paper, we combine methods from both fields in order to guarantee that a candidate strategy learned through machine learning provably satisfies temporal logic specifications.

At first, we learn a randomized strategy\footnote{Also referred to as stochastic strategy or policy.} via \acp{RNN}~\cite{hochreiter1997long} and data stemming from knowledge of the underlying structure of \acp{POMDP}.
We refer to the resulting trained \ac{RNN} as the \emph{strategy network}.
\acp{RNN} are a good candidate for learning a strategy because they can successfully represent temporal dynamic behavior~\cite{pascanu2013construct}.

Secondly, we extract a concrete (memoryless randomized) candidate strategy from the \ac{RNN} and use it directly on a given \ac{POMDP}, resulting in the \ac{DTMC} induced by the \ac{POMDP} and the strategy.
Formal verification reveals whether specifications are satisfied or not.
In the latter case, we generate a so-called counterexample~\cite{DBLP:journals/tcs/WimmerJAKB14}, which points to parts of the \ac{DTMC} (and by extension of the \ac{POMDP}), that are critical for the specification.
For those critical parts, we use a \ac{LP} approach that locally improves strategy choices (without any guarantees on the global behavior).
From the improved strategy, we generate new data to retrain the \ac{RNN}.
We iterate that procedure until the strategy network yields satisfactory results.

While the strategies are memoryless, allowing for randomization over possible choices --~relaxing determinism~-- is often sufficient to capture necessary variability in decision-making.
The intuition is that \emph{deterministic} choices at a certain state may need to vary depending on previous decisions, thereby trading off memory.
However, randomization in combination with finite memory may supersede infinite memory even more for many cases~\cite{amato2010optimizing,junges2018finite}.
We encode finite memory directly into a POMDP by extending its state space.
We can then directly apply our method to create \emph{\acp{FSC}}~\cite{meuleau1999learning}.

As previously discussed, the investigated problem is undecidable for \acp{POMDP}~\cite{MadaniHC99} and therefore the approach is naturally incomplete. Soundness is provided, as verification yields hard guarantees on the quality of a  strategy.

\paragraph*{Related Work.} 
We list relevant works in addition to the ones already mentioned.
\cite{wierstra2007solving} is the first to employ a \ac{LSTM} architecture to learn (finite-memory) strategies for \acp{POMDP}.
%They are the first to show that \acp{LSTM} are able to learn strategies which leverage events that lie arbitrarily far in the past.
\cite{mnih2015human} develops a \ac{NN}-based Q-learning algorithm (called Deep Q-learning) to play video games straight from video frames, under partial observability.
%The authors deal with partial observability using the last 4 most recent frames as inputs.
%Although the authors do not use \acp{RNN}, the memory problem is solved using this replay trick.
\cite{hausknecht2015deep} uses an \ac{LSTM} cell to enhance the algorithm with memory.
%The authors also introduce more partial observability by obscuring frames or parts of the frames before processing.
%Still, the algorithm learns quickly to adapt and  synthesize powerful strategies.
%Notable point-based methods are PBVI~\cite{pineau2003point}, HSVI~\cite{smith2004heuristic}, Perseus~\cite{spaan2005perseus}, and SARSOP~\cite{kurniawati2008sarsop}. 
While these approaches yield good performance, they do not provide formal guarantees on strategies and cannot incorporate temporal logic specifications.
\cite{chatterjee2015qualitative} studies verification problems for \acp{POMDP} with temporal logic specifications on a theoretical level without a connection to machine learning.
%The proposed approach differs through its iterative procedure to find satisfying strategies that the heuristics used in \cite{chatterjee2015qualitative} may not capture.
%Finally, \cite{carr2018human}, which synthesizes strategies for \ac{POMDP} with specifications using a human-in-the-loop procedure, rather than an \ac{RNN}, to resolve the partial observability and non-determinism.
%These methods approximate either the belief space, the value functions, the strategy, or a combination of all three.

%POMCP~\cite{silver2010monte} uses a Monte-Carlo sampling method 
%which simulates the POMDP, analyzes with the so-called ``underlying difficulty of the POMDP" rather than the size of the state space 
%to find approximate solutions to problems with millions of states.

A different research area addresses the (direct) verification of \acp{NN}, as opposed to the model-based verification approach we pursue, see for instance~\cite{katz2017reluplex}.
%For instance, the output of 
% that neural networks do not perform differently when the input is slightly perturbed.
%The core idea is to encode an \ac{NN} output as a set of constraints to capture potential perturbations or safety violations and use SMT-solvers~\cite{katz2017reluplex,ehlers2017formal,DBLP:conf/cav/HuangKWW17} or mixed integer programming~\cite{cheng2017maximum} to obtain certificates.

%\paragraph*{Structure of the paper: }
%The rest of the paper is structured as follows.
%After formal foundations on \ac{POMDP} in Sect.~\ref{sec:preliminaries}, Sect.~\ref{sec:method} describes the synthesis procedure.
%In Sect.~\ref{sec:DRNN}, we detail how we sample the strategy using a \ac{RNN}.
%We demonstrate the applicability of the proposed approach using a selection of temporal logic examples as well as comparing to well-known benchmarks~\cite{smith2004heuristic} in Sect.~\ref{sec:experiments}.

% !TeX spellcheck = en_US
\acresetall

\section{Preliminaries}
\label{sec:preliminaries}
\noindent A \emph{probability distribution} over a finite or countably infinite set $\distDom$
is a function $\distFunc\colon\distDom\rightarrow\Ireal$ with $\sum_{\distDomElem\in\distDom}\distFunc(\distDomElem)=\distFunc(\distDom)=1$.
The set of all distributions on $\distDom$ is $\Distr(\distDom)$. The support of a distribution $\distFunc$ is
$\supp(\distFunc) = \{x\in\distDom\,|\,\distFunc(x)>0\}$.
%A distribution is \emph{Dirac} if $|\!\supp(\distFunc)| = 1$.
%\subsection{Probabilistic Models}
%\label{ssec:prob_models}
%\begin{definition}[MDP]
%  \label{def:pmdp}
\paragraph{(PO)MDPs.}
  A \emph{\ac{MDP}} $\mdp$ is a tuple $\MdpInit$ with
  a finite (or countably infinite) set $\states$ of \emph{states},
%  \emph{initial state} $\sinit\in\states$,
  a finite set $\Act$ of \emph{actions},
  and a \emph{transition function} $\probmdp\colon \states\times\Act\rightarrow\Distr(\states)$.
We use a reward function $\rew\colon\states\times\act\rightarrow\R$.
%  
%  
%  \states\rightarrow [0,1]$ such that
%  $\forall s\in S\ \forall\act\in\Act: \sum_{s'\in S}\probmdp(s,\act,s')\in\{0,1\}$.
%\end{definition}
%
%The \emph{available actions} in $s\in\states$ are $\Act(s)=\{\act\in\Act\mid \exists s'\in\states: \probmdp(s,\act,s') > 0\}$.
%We assume that MDP $\mdp$ contains no deadlock states, \ie $\Act(s)\neq\emptyset$ for all $s\in\states$.
A finite \emph{path} $\pi$ of an \ac{MDP} $\mdp$ is a sequence of states and actions; $\last{\pi}$ is the last state of $\pi$.
%
%(in)finite sequence of 
%$\pi = s_0\xrightarrow{\act_0}s_1\xrightarrow{\act_1}\cdots$,
%where $s_i\in\states$, $\act_i\in\Act(s_i)$, and $\probmdp(s_i,\act_i,s_{i+1}) > 0$ for all $i\in\N$.
%
The set of finite paths of $\mdp$ is $\pathsfin^{\mdp}$.
A \emph{\ac{DTMC}} is an \ac{MDP} with $|\Act(s)|=1$ for all $s\in\states$.
%For an \ac{DTMC} $\dtmc$, we may omit the actions and use the notation $\dtmcInit$ with a transition function $\probdtmc$
%of the form $\probdtmc\colon\states\times\states\rightarrow \R$.

%\paragraph{Strategies.}
%To resolve the action choices in \acp{MDP}, \emph{strategies}
%%\footnote{Also referred to as policies or schedulers.} 
%determine distributions over actions, potentially based on the \emph{history} of the current path.
%This decision may be based on the \emph{history} of the current path.
%
%\begin{definition}[Strategy]
%  \label{def:strategy}
  A \emph{strategy} $\sched$ for and MDP $\mdp$ is a function $\sched\colon \pathsfin^{\mdp}\to\Distr(\Act)$
  with $\supp\bigl(\sched(\pi)\bigr) \subseteq \Act\bigl(\last{\pi}\bigr)$ for all $\pi\in \pathsfin^{\mdp}$.
%  The set of all strategies for $\mdp$ is $\Sched^{\mdp}$.
%\end{definition}
%
A strategy $\sched$ is \emph{memoryless} if $\last{\pi}=\last{\pi'}$ implies $\sched(\pi)=\sched(\pi')$ for all $\pi,\pi'\in\pathsfin^{\mdp}$.
%It is \emph{deterministic} if $\sched(\pi)$ is a Dirac distribution for all $\pi\in\pathsfin^{\mdp}$.
%A strategy that is not deterministic is \emph{randomized}.
%
%A strategy resolves all nondeterministic choices, yielding an \emph{induced Markov chain}.
%For a DTMC and a fixed initial state $\sinit$ (or a probability distribution over the initial states),
%a \emph{probability measure} over the set of infinite paths is defined~\cite{BK08}.
%
\begin{definition}[Induced Markov Chain]
  \label{def:induced_dtmc}
 For an \ac{MDP} $\MdpInit$ and a strategy $\sched\in\Sched^{\mdp}$, the \ac{DTMC} induced by $\mdp$ and $\sched$ is given by $\DTMCgnd = (\pathsfin^{\mdp},\probdtmc^{\sched})$ where:
  \[
    \probdtmc^{\sched}(\pi,\pi') = \begin{cases}
        \probmdp(\last{\pi},\act,s')\cdot\sched(\pi)(\act) & \text{if $\pi' = \pi\act s'$,} \\
        0 & \text{otherwise.}
      \end{cases}
  \]
\end{definition}
%\spc{Nils: define probability measure}
%
%\subsection{Partial Observability}
%\label{ssec:partial_obs}
%
\begin{definition}[POMDP]
  \label{def:pomdp}
  A \emph{\ac{POMDP}} is a tuple $\PomdpInit$, with $\MdpInit$ the \emph{underlying MDP of $\pomdp$}, $\ObsSym$ a finite set of observations and $\ObsFun\colon\states\rightarrow\ObsSym$ the \emph{observation function}.
\end{definition}
The set of all finite observation-action sequences for a \ac{POMDP} $\pomdp$ is denoted by $\obsSeqFin^\pomdp$. 
%An observation-based strategy selects actions based on observations along a path and the past actions. 
%%
\begin{definition}[POMDP Strategy]
  \label{def:obsstrategy}
  An \emph{observation-based strategy} for a \ac{POMDP} $\pomdp$ is a function $\osched\colon\obsSeqFin^\pomdp\rightarrow\Distr(\Act)$
  such that $\supp\bigl(\osched(\ObsFun(\pi))\bigr) \subseteq \Act\bigl(\last{\pi}\bigr)$ for all $\pi\in \pathsfin^{\mdp}$.
  $\oSched^\pomdp_\obs$ is the set of observation-based strategies for $\pomdp$.
\end{definition}
A \emph{memoryless} observation-based strategy $\osched\in\oSched^\pomdp_\obs$ is analogous to a memoryless \ac{MDP} strategy, formally we simplify to $\osched\colon Z\rightarrow\Distr(\Act)$, \ie we decide based on the current observation only.
Similarly, \ac{POMDP} together with a strategy yields an induced \ac{DTMC} as in Def.~\ref{def:induced_dtmc}, resolving all nondeterminism and partial observability.
A general \ac{POMDP} strategy can be represented by \emph{infinite-state controllers}.
Strategies are often restricted to finite memory; this amounts to using \acp{FSC}~\cite{meuleau1999learning}.
%The set $FSC_k^\mathcal{M}$ denotes the set of \acp{FSC} with $k$ memory nodes, called $k$-FSCs.
%
%$\osched^o\in oSched^\pomdp_\obs$ considers only the last observation in the sequence for providing a distribution, $\osched^o:\ObsFun(\last\pi)\rightarrow \Distr(\Act)$.

%
\begin{definition}[\ac{FSC}]
	\label{def:FSCstrategy}
	A $k$-\ac{FSC} for a \ac{POMDP} is a tuple $\mathcal{A} = (N,n_I,\gamma, \delta)$ where $N$ is a \emph{finite set} of $k$ memory nodes, $n_I \in N$ is the initial memory node, $\gamma$ is the action mapping $\gamma\colon N \times \ObsSym \rightarrow \Distr(\Act)$ and $\delta$ is the memory update $\delta\colon N\times \ObsSym \times \Act \rightarrow N	$.
%	The set $FSC_k^\mathcal{M}$ denotes the set of \acp{FSC} with $k$ memory nodes, called $k$-FSCs.
	Let $\osched_{\mathcal{A}} \in \oSched^\pomdp_\obs$ denote the observation-based strategy represented by the \ac{FSC} $\mathcal{A}$.
\end{definition}
%
%A \emph{$k$-\ac{FSC}} $\mathcal{A}$ has $k$ memory nodes which determine the action choice of the strategy.
The product $\pomdp\times\mathcal{A}$ of a POMDP and a $k$-\ac{FSC} yields a (larger) ``flat'' POMDP where the memory update is directly encoded into the state space~\cite{junges2018finite}.
The action mapping $\gamma$ is left out of the product.
A memoryless strategy $\osched\in\oSched_\obs^{\pomdp\times\mathcal{A}}$ then determines the action mapping and can be projected to the finite-memory strategy $\osched_{\mathcal{A}} \in \oSched^\pomdp_\obs$.
%From a node $n$ and the observation $\obs$ in the current state of the \ac{POMDP}, the next action $a \in \Act$ is chosen from the action mapping $\gamma(n,\obs)$. The successor node of the \ac{FSC} is updated via $\delta(n,\obs,a)$. 
%
%
\paragraph{Specifications.}
\label{ssec:spec}
We consider \ac{LTL} properties~\cite{Pnueli77}. 
For a set of atomic propositions $AP$, which are either satisfied or violated by a state, and $a\in AP$, the set of LTL formulas is given by:
\[
    \Psi\coloncolonequals a \ |\  (\Psi\land\Psi)\ |\ \neg\Psi\ |\ \Next\Psi\ |\ \Always\Psi\ |\ (\Psi\Until\Psi)\,.
\]
Intuitively, a path $\pi$ satisfies the proposition $a$ if its first state does; $(\ltlformula_1\land\ltlformula_2)$ is satisfied,
if $\pi$ satisfies both $\ltlformula_1$ and $\ltlformula_2$; $\neg\ltlformula$ is true on $\pi$ if $\ltlformula$ is not satisfied.
The formula $\Next\ltlformula$ holds on $\pi$ if the subpath starting at the second state of $\pi$ satisfies $\ltlformula$.
The path $\pi$ satisfies $\Always\ltlformula$ if all suffixes of $\pi$ satisfy $\ltlformula$. Finally, $\pi$ satisfies
$(\ltlformula_1\Until\ltlformula_2)$ if there is a suffix of $\pi$ that satisfies $\ltlformula_2$ and all longer suffixes satisfy
$\ltlformula_1$. $\Finally\ltlformula$ abbreviates $(\mathrm{true}\Until\ltlformula)$.
%and requires
%that $\pi$ needs to have a suffix that satisfies $\ltlformula$.

For POMDPs, one wants to synthesize a strategy such that the probability of satisfying an LTL-property respects a given bound, denoted $\reachPropSymbol = \p_{\sim \lambda}(\ltlformula)$ for ${\sim}\in\{ {<}, {\leq},{\geq},{>}\}$ and $\lambda\in[0,1]$.
In addition, \emph{undiscounted expected reward properties} 
$\reachPropSymbol=\Ex_{\sim \lambda}(\Finally a)$ require that the expected accumulated cost until reaching a state satisfying $a$ respects $\lambda\in\R_{\geq 0}$.
%\spc{In the EC I think $\lambda$ isn't restricted to $\lambda\in[0,1]$}

If $\reachPropSymbol$ (either LTL or expected reward specification) is satisfied in a (PO)MDP $\pomdp$ under $\sched$, we write $\pomdp^\sched\models\reachPropSymbol$, that is, the specification is satisfied in the induced \ac{DTMC}, see Def.~\ref{def:induced_dtmc}.
%
%
%\ralf{Issue regarding memory: For checking LTL-properties one typically needs memory also for MDPs. This memory is provided in form of a Rabin/Muller/... automaton.
%    By using the product automaton of the (PO)MPD and the Rabin automaton, the LTL property is reduced to repeated reachability. For MDPs then a memoryless
%    strategy suffices, for POMDPs one might need additional memory due to the restricted observability. In my opinion, leaving out/restricting the
%    memory required for encoding the property make little sense, restricting the memory to cope with partial observability is unavoidable and therefore ok.
%    Could you please clarify how we handle this?
While determining an appropriate strategy is still efficient for MDPs, this problem is in general undecidable for POMDPs~\cite{ChatterjeeCT16}.
In particular, for MDPs, to check the satisfaction of a general LTL specification one needs memory.
Typically, tools like \tool{PRISM}~\cite{KNP11} compute the product of the MDP and a deterministic Rabin automaton. 
In this product, reachability of so-called accepting end-components ensures the satisfaction of the LTL property. 
This reachability probability can be determined in polynomial time.
\tool{PRISM-POMDP}~\cite{NPZ17} handles the problem similarly for POMDPs, but note that a strategy needs memory not only for the LTL specification but also for observation dependencies.

Finally, given a (candidate) strategy $\sched$, checking whether $\pomdp^\sched\models\varphi$ holds can be done both for MDPs and POMDPs in polynomial time.
For more details we refer to \cite{BK08}.

\section{Synthesis Procedure}
\label{sec:method}

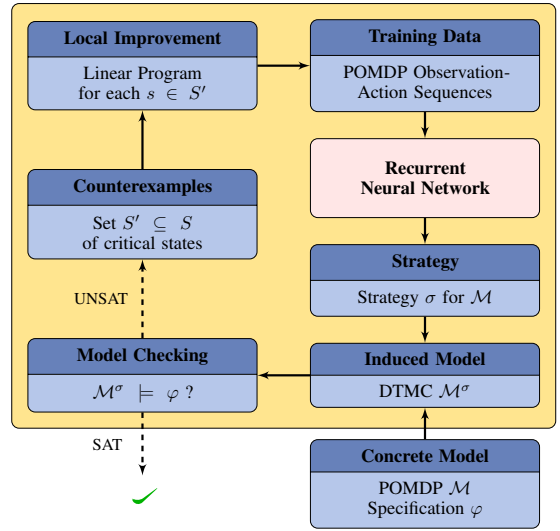
\begin{figure}[!ht]
	\centering
	{\scalebox{0.7}{\definecolor{bg}{HTML}{ddeedd}
\definecolor{comp}{HTML}{c2d4dd}
\definecolor{impl}{HTML}{b0aac0}
\definecolor{ligb}{HTML}{5E7FC6}
\definecolor{bodybl}{HTML}{85A1DC}
\definecolor{headbl}{HTML}{264C9C}
\definecolor{bgyel}{HTML}{FFDC6B}
\definecolor{bodyyel}{HTML}{FFE58F}
\definecolor{headyel}{HTML}{E9BB25}
\centering
\begin{tikzpicture}[every node/.style={draw, text centered, shape=rectangle, rounded corners, text width=4cm, minimum height=1.5cm, inner sep=5pt}]
%\tikzstyle{outer}= [draw, text centered, shape=rectangle, text width=2cm, minimum height=1cm]
%\tikzstyle{inner}=[draw, text centered, shape=rectangle, rounded corners, text width=3.8cm, minimum height=1.1cm, inner sep=5pt]
\tikzset{
	splitnode/.style={
		rectangle split,
		rectangle split parts=2,
		rectangle split part fill={headbl!70,bodybl!60}
	}
}
\tikzset{
	normalnode/.style={
		fill=red!10
	}
}
\tikzstyle{split}=[rectangle split,rectangle split parts=2]

\node[splitnode](lp) {\textbf{Local Improvement} \nodepart{second} Linear Program for each $s\in S'$};
\node[splitnode,right=1.0cm of lp] (newsamp) {\textbf{Training Data} \nodepart{second} POMDP Observation-Action Sequences};
\node[normalnode, below=0.5cm of newsamp] (drnn) {\textbf{Recurrent\\Neural Network}};
\node[splitnode, below=0.5cm of drnn] (strategy) {\textbf{Strategy} \nodepart{second} Strategy $\osched$ for $\pomdp$};
\node[splitnode, below=0.5cm of strategy] (dtmc) {\textbf{Induced Model} \nodepart{second} DTMC $\pomdp^\osched$};

\node[splitnode, left=1.0cm of dtmc] (model) {\textbf{Model Checking} \nodepart{second} $\pomdp^\sigma\models\reachPropSymbol$~?};
\node[splitnode,above=1.5cm of model](count) {\textbf{Counterexamples} \nodepart{second} Set $S'\subseteq S$ of critical states};

%\node[text width=2.5cm,minimum height=1.3cm,draw=none,right=1.7cm of newsamp] (drnn) {\textbf{Recurrent\\Neural Network}\vspace{6pt} \includegraphics[width=0.8\textwidth]{pics/layer_icon}};

\begin{scope}[on background layer]
\node [fit = (current bounding box), inner sep = .3cm,  fill=bodyyel] (box) {};
\end{scope}

%\node[splitnode, right=2.65cm of drnn] (trajsamp) {\textbf{Trajectory Samples} \nodepart{second} For entire problem class};

%\node[normalnode, below=1cm of model, minimum height=1cm, text width=2cm] (spec) {\textbf{Specification}};

\node[splitnode, below=0.6cm of dtmc] (pomdp) {\textbf{Concrete Model} \nodepart{second}POMDP $\pomdp$ \\ Specification $\reachPropSymbol$ };

\node[draw=none,below=1.2cm of model,minimum height=0.1cm,text width =0.5cm] (safe) {\color{green!70!black}\Large\checkmark};

%%arrows
%\draw ([yshift=-2ex] testing.west) edge[-latex', very thick] ([yshift=-2ex] system.east);
%\draw ([yshift=2ex] system.east) edge[-latex', very thick] ([yshift=2ex] testing.west);

\path [line,-latex', very thick, dashed] (model) --node[draw=none,left,text width=1.2cm]{\small UNSAT } (count);
\path [line,-latex', very thick, dashed] (model) --node[draw=none,left=0.5cm,text width=0.25cm,minimum height=0.05cm]{\small SAT} (safe);

\draw (count) edge[-latex', very thick] (lp);

\draw (lp) edge[-latex', very thick] (newsamp);

\draw (newsamp) edge[-latex', very thick] (drnn);

\draw (drnn.south) edge[-latex', very thick] (strategy.north);

\draw (strategy.south) edge[-latex', very thick] (dtmc.north);

\draw (dtmc.west) edge[-latex', very thick] (model.east);

\draw (pomdp) edge[-latex', very thick] (dtmc);
%\draw (pomdp) edge[-latex', very thick] (trajsamp);
%\draw (trajsamp) edge[-latex', very thick] (drnn);
\end{tikzpicture}}}
	\caption{Flowchart of the \ac{RNN}-based refinement loop}
	\label{fig:flowchart}
\end{figure}

\begin{mdframed}[backgroundcolor=gray!30]
\textbf{Formal Problem Statement.} For a \ac{POMDP} $\pomdp$ and a specification $\reachPropSymbol$, where either $\reachPropSymbol=\p_{\sim \lambda}(\ltlformula)$ with $\ltlformula$ an LTL formula, or $\reachPropSymbol=\Ex_{\sim\lambda}(\Finally a)$,
the problem is to determine a (finite-memory) strategy $\osched\in\oSched^\pomdp_\obs$ such that $\osched\models\reachPropSymbol$.
\end{mdframed}
If such a strategy does not exist, the problem is infeasible.

\paragraph{Outline.} The workflow of the proposed approach is illustrated in Fig.~\ref{fig:flowchart}: We \emph{train} an \ac{RNN} using observation-action sequences generated from an initial strategy as discussed in Sect.~\ref{ssec:DRNN}.
The trained \emph{strategy network} represents an observation-based strategy, taking as input an observation-action sequence and returning a distribution over actions, see Def~\ref{def:obsstrategy}.
For a POMDP $\pomdp$, we use the output of the strategy network in order to resolve nondeterminism.
The strategy network is thereby used to extract a \emph{memoryless strategy} $\osched\in\oSched^\pomdp$ and as a result we obtain the induced \ac{DTMC} $\pomdp^\osched$.
%By sampling the \ac{RNN} for the relevant observations (or sequences)\as{There is no such thing as sampling an RNN. How about 'we take the output' and also describe what the output is. Also, we input a sequence or an observation? We should be precise about it.}, we obtain a memoryless (or finite-memory strategy) $\osched\in\oSched^\pomdp_{\obs}$ for the \ac{POMDP} $\pomdp$.
%We apply that strategy to the \ac{POMDP} and obtain the induced MC $\pomdp^\osched$. 
\emph{Model checking} of this induced \ac{DTMC} evaluates whether the specification $\reachPropSymbol$ is satisfied or not for the extracted strategy.
In the former case, the synthesis procedure is finished.
The extraction and evaluation is explained in Sect.~\ref{sec:extract_evaluate}.

If the specification is not satisfied, we obtain a \emph{counterexample} highlighting critical states of the \ac{POMDP}.
We employ a \ac{LP} approach that locally \emph{improves} action choices of the current strategy at these critical states, see Sect.~\ref{ssec:drnn_improve}.
Afterwards, we retrain the \ac{RNN} by generating new observation-action sequences obtained from the new strategy.
We iterate this procedure until the specification is satisfied or a fixed iteration threshold is reached.
For cases where we need to further improve, we use domain knowledge to create a specific memory-update function of a $k$-\ac{FSC} $\fsc$, see Def.~\ref{def:FSCstrategy}.
Then, we compute the product $\pomdp'=\pomdp\times\fsc$.
We iterate our method with $\pomdp'$ as starting point and thereby determine a concrete $k$-\ac{FSC} including the action mapping.

%\as{It is unclear what 'domain knowledge' means at this stage. Can we be more descriptive or just remove this completely?}
%A \textbf{finite-memory strategy} for $\pomdp$ is determined by projecting the resulting memoryless strategy $\osched\in \oSched^{\pomdp'}$ to the state space of $\pomdp$. 

%\subsection{Training a \ac{RNN} as a strategy network}
%% SECTION 3.2 STARTS HERE
\subsection{Learning Strategies with \acp{RNN}}
\label{ssec:DRNN}
Optimization methods to approximate strategies fall within the policy gradient class of algorithms, specific to reinforcement learning~\cite{sutton2000policy}.
In this setting, the strategy is parametrized and updated by performing gradient ascent on the error function (typically chosen to maximize the discounted reward).
%In this setting, the strategy is parametrized by a set of parameters $\vtheta \in \Theta$\nils{$\Theta$ has not been introduced}, estimated using the gradient of a performance measure $J(\vtheta)$, where $J(\vtheta)$ aims to maximize the average (or discounted) reward.
%Therefore, the parameter update can be obtained by performing gradient ascent on $J(\vtheta)$, as follows: $\vtheta_{t+1} = \vtheta_{t} + \eta\cdot\nabla(J(\vtheta_t))$, where the learning rate $\eta$ controls the contribution of the gradient in the update.
Policy gradient algorithms are generally used to map observations to actions, and are not well suited for \acp{POMDP} due to their inability to cope with arbitrary memory.
To overcome this weakness, we design our method to make explicit use of memory, using \acp{RNN}, which are a family of neural networks designed to learn dependencies in sequential data.
They leverage an internal state to process and store information between sequential steps, thus simulating memory.

%Moreover, \acp{RNN} are differentiable end-to-end and make explicit use of gradient information to update their parameters (a procedure called back propagation through time).
%Storing long-term dependencies in the internal state has been a problem for \acp{RNN} due to their entangled structure.
%Deciding which information to retain, which to discard, and which to forward in one step is difficult for many tasks.
%Moreover, training \acp{RNN} has suffered from a phenomenon called vanishing (or exploding) gradient: when propagating the error back through time, the gradient values either diminish or increase very fast.

%Two architectures were designed on purpose to deal with the problems mentioned above: \ac{LSTM}~\cite{hochreiter1997long} and, later, gated \acp{RNN}~\cite{chung2014empirical}.
%They leverage the concept of gates to separate the steps for discarding, storing, and processing information, thus allowing to explicitly learn parameters for each operation and easing the entanglement.
%\acp{LSTM} and gated \acp{RNN} can be trained in a supervised manner and are heavily used in tasks which require time dependencies, such as speech recognition or language translation.\nj{We need to shorten the last two paragraphs a bit, other than that they are nice.}

\paragraph{Constructing the Strategy Network.}
We use the \ac{LSTM} architecture~\cite{hochreiter1997long} in a similar fashion to policy gradient methods and model the output as a probability distribution on the action space (described formally by $\RNNfun\colon \obsSeqFin^\pomdp \rightarrow \Distr(\Act) )$.
Having stochastic output units, we avoid computing gradients on the internal belief states, as it is, for example, done in~\cite{meuleau1999learning}.
Using back propagation through time, we can update the strategy during training.
Thus, for a given observation-action sequence from $\obsSeqFin^\pomdp$, the model learns a strategy $\RNNfun\in \oSched^\pomdp_\obs$.
The output is a discrete probability distribution over the actions $\Act$, represented using a final softmax layer.%\nils{Alex, is this really a discrete distribution stored in the RNN?\rw{Discrete since it assigns a proability to a discrete set of elements.}}

\paragraph{RNN Training.}
We train the \ac{RNN} using a slightly modified version of sampling re-usable trajectories~\cite{kearns2000approximate}.
In particular, for a \ac{POMDP} $\PomdpInit$ and a specification $\varphi$, instead of randomly generating observation sequences, we first compute a strategy $\sched\in\Sched^\mdp$ of the underlying MDP $\mdp$ that satisfies $\varphi$.
Then we sample uniformly over all states of the \ac{MDP} and generate finite paths (of a fixed maximal length) from $\pathsfin^{\mdp^{\sched}}$ of the induced \ac{DTMC} $\mdp^\sched$, thereby creating multiple trajectory trees.
%
%from the initial set of states, it builds multiple trajectory trees based on the possible set of initial states and limits each tree to only those nodes defined by 
%Value iteration with a fixed discount-rate can find such a policy in polynomial time \cite{littman1995complexity}.
%These trajectory trees can be formally described by the set of paths .
For each finite path $\pi\in \pathsfin^{\mdp^{\sched}}$, we generate one possible observation-action sequence $\pi_\obs\in \obsSeqFin^\pomdp$ such that $\pi=\obs_0,\act_0,\ldots,\act_{n-1},\obs_n$ with $\obs_i=\ObsFun(\pi[i])$, where $\pi[i]$ denotes the $i$-th state of $\pi$ for all $1\leq i\leq n$.
%
%
%When sampling, selecting one of the trees and following it to a leaf, which forms either at the fixed horizon or a deadlock, gives a finite path $\pi$, from which we generate an observation-action sequence $\ObsFun(\pi) \in \obsSeqFin$.
%For each path sampled $\ObsFun(\pi)$, we remove the last state in the path and 
%
We form the training set $\mathcal{D}$ from a (problem specific) number of $m$ observation-action sequences with observations as input and actions as output labels.
%
%The \ac{RNN} is trained on this dataset $\mathcal{D}$ with the observations $\lbrace \ObsFun(s_0),\cdots,\ObsFun(s_{n-1}) \rbrace$ as the input and each action $\lbrace a_0,\cdots,a_{n-1}\rbrace$ as the output labels.
Both input and output sets were processed using one-hot-encoding.
To fit the \ac{RNN} model, we use the Adam optimizer~\cite{kingma2014adam} with a cross-entropy error function.
%\nils{as I have no clue what the last sentences mean, Alex, Steve, please check if all makes sense.}
%\alex{I checked it and it is accurate.}

\paragraph{Sampling Large Environments.}
In a \ac{POMDP} $\pomdp$ with a large state space ($|\states|>10^5$), computing the underlying \ac{MDP} strategy $\sigma\in \Sigma^\mdp$ affects the performance of the procedure.
In such cases, we restrict the sampling to a smaller environment that shares the observation $\ObsSym$ and action spaces $\Act$ with $\pomdp$.
For example, consider a gridworld scenario with a moving obstacle that has the same underlying probabilistic movement for different problem sizes; such a framework can provide a similar dataset regardless of the size of the grid.

\subsection{Strategy Extraction and Evaluation}\label{sec:extract_evaluate}
We first describe how to extract a memoryless strategy from the strategy network for a specific POMDP, then we formalize the extension to \acp{FSC} to account for finite memory.
Afterwards, we shortly explain how to evaluate the resulting strategies.
%\paragraph{Extracting a finite memory strategy from an RNN.}\nils{finite memory or memoryless?}
%Due to the sensitivity to initial conditions and additional complexity introduced by quantization, directly verifying a strategy network remains difficult, inconsistent and inefficient \cite{kolen1994fool}.
%\as{I never mention any reasons. I think we should do it here if it is needed.}
%\spc{I had a go at this, I think one line summarizing these reasons are sufficient but please feel free to change for correctness/completeness \nj{I think that does not belong here, we can sufficiently describe in the related work paragraph.}}

Given a \ac{POMDP} $\pomdp$, we use the trained strategy network $\RNNfun\colon \obsSeqFin^\pomdp \rightarrow \Distr(\Act)$ directly as observation-based strategy.
Note that the \ac{RNN} is inherently a predictor for the distribution over actions and will not always deliver the same output for one input.
While we always use the first prediction we obtain, one may also sample several predictions and take the average of the output distributions.
%\ralf{What does that sentense mean? Average over what?}
%\steve{}
%
%At an individual observation $(\obs\in \ObsSym)$, we map the network's prediction for the output distribution over actions $RNN(\obs) = \Distr(\Act)$ to a memoryless strategy: $\osched(\obs) = RNN(\obs)$. 
%Continuing this process for the set of all possible observations in the environment creates a randomized memoryless strategy $\left(\forall \obs\in\ObsSym ~ \osched(\obs) = RNN(\obs)\right)$.
% ( $\osched\colon\obsSeqFin^\pomdp\rightarrow\Distr(\Act)$ where $\obsSeqFin^\pomdp$ is the set of all single-state observation sequences).

\paragraph{Extension to \ac{FSC}s.}
As mentioned before, LTL specifications as well as observation-dependencies in \ac{POMDP}s require memory. 
Consider therefore a general \ac{FSC} $\mathcal{A} = (N,n_I,\gamma,\delta)$ as in Def.~\ref{def:FSCstrategy}.
%
%, directly extracting the memory update function $\delta$ is difficult, inconsistent and inefficient due to the additional complexity introduced by quantization \cite{kolen1994fool}.
%However, these difficulties do not prevent us from constructing an \ac{FSC} $\mathcal{A}$ using this procedure.
We first predefine the memory update function $\delta$ in a problem-specific way, for instance, $\delta$ changes the memory node when an observation is repeated.
Consider observation sequence $\pi_\obs\in \obsSeqFin^\pomdp$ with $\pi_\obs=\obs_0,\act_0,\ldots,\obs_n$.
Assume, the FSC is in memory node $n_{k}\in N$ at position $i$ of $\pi_\obs$. We define $\delta(n_k,\obs_i,\act_i)=n_{k+1}$, if $\pi_\obs[i]=(\obs_i,\act_i)$, and there exists a $j<i$ such that $\pi_\obs[j]=(\obs_j,\act_j)$ with $\obs_i=\obs_j$. 
Similarly, we account for specific memory choices akin to the relevant LTL specification.
%
%
%for $\pi(s_i) \in \obsSeqFin^\pomdp$, the memory update function is  $\left(\ObsFun(\pi(s_i))\equiv\ObsFun(next(\pi(s_{i}) )) \right) \rightarrow \delta(n_{k}) = (n_{k}+1)\mod |N|$ with probability 1 and elsewhere $\delta(n_{k}) = n_k$.
%Note that such a deterministic update rule is problem and environment dependent.

Once $\delta$ has been defined, we compute a product \ac{POMDP} $\pomdp\times\mathcal{A}$ which creates a state space over $S \times N$.
The training process is similar to the method outlined above but instead of generating observation-action sequences from $\obsSeqFin^\pomdp$, we generate observation-node-action sequences $(\obs_0,n_0),\act_0,\ldots,\act_{n-1},(\obs_n,n_n)$ from $\obsSeqFin^{\pomdp\times\mathcal{A}}$.
%
%$\ObsFun(\pi) \times \delta(\pi) \in \obsSeqFin \times N$.
%The training input data becomes a sequence of these tuples $(\obs_i,n_i)^*\in \obsSeqFin \times N$: $\mathcal{D}_{FSC}$ has an input of $\lbrace (\ObsFun(s_0),n_0),\cdots,(\ObsFun(s_{m-1}),n_{m-1}) \rbrace$.
In this case, the \ac{RNN} is learning the mapping of observation and memory node to the distribution over actions as an \ac{FSC} strategy network: $\RNNfun_{\mathrm{FSC}}\colon \obsSeqFin^{\pomdp\times\mathcal{A}} \times N \rightarrow \Distr(\Act)$

In order to extract the memoryless \ac{FSC} $\mathcal{A}$ from the \ac{FSC} strategy network $\RNNfun_{\mathrm{FSC}}$, we collect the predicted distributions across the product set of all possible observations $z\in \ObsSym$ and all possible memory nodes $n \in N$.
%In particular, we get $\RNNfun_{\mathrm{FSC}}\colon \ObsSym\times N\rightarrow \Distr(\Act)$.
From this prediction, the \ac{FSC} $\mathcal{A}$ is constructed from the action mapping $\gamma(\obs,n)=\RNNfun_{\mathrm{FSC}}(\obs,n) $ and the predefined memory update function $\delta$.

\paragraph{Evaluation.}
We assume that for \ac{POMDP} $\PomdpInit$ and specification $\varphi$, we have a finite-memory observation-based strategy $\osched\in\oSched^\pomdp$ as described above. 
We use the strategy $\osched$ to resolve all nondeterminism in $\pomdp$, resulting in the induced \ac{DTMC} $\pomdp^\osched$, see Def.~\ref{def:induced_dtmc}.
For this \ac{DTMC}, we apply model checking, which in polynomial time reveals whether $\pomdp^\osched\models\varphi$.
For the fixed strategy $\osched$ we extracted from the strategy network, this provides hard guarantees about the quality of $\osched$ regarding $\varphi$.
As mentioned before, this strategy is only a prediction obtained from the \ac{RNN} -- so the guarantees necessarily do not directly carry over to the strategy network.

\subsection{Improving the Represented Strategy}
\label{ssec:drnn_improve}

%As outlined in Def.~\ref{def:induced_dtmc}, we apply the strategy $\osched$ to the concrete POMDP scenario $\pomdp$, which resolves all non-determinism and partially observability to create an induced MC $\pomdp^\osched$.
%We can efficiently verify this MC against the specification $\reachPropSymbol$ using probabilistic model checking (PMC) tools.
%Consider the specification such as \textit{the probability reaching a goal area without bumping into obstacles is above a certain threshold} (see Sect.~\ref{sec:experiments}).
%If for this DTMC, the specification is satisfied, then the computed strategy provably induces this exact specification.
%If the strategy does not satisfy the specification, we generate specific counterexamples based on a linear programming formulation for selecting problematic states $s'\in S$.

We describe how we compute a \emph{local improvement} for a strategy that does not satisfy the specification.
In particular, we have \ac{POMDP} $\PomdpInit$, specification $\varphi$, and the strategy $\osched\in\oSched^\pomdp$ with $\pomdp^\osched\not\models\varphi$.
We now create diagnostic information on why the specification is not satisfied.

First, without loss of generality, we assume $\varphi=\p_{\leq \lambda}(\ltlformula)$.
Let $\osched(\obs)(\act)$ denote the probability of choosing action $\act\in\Act$ upon observation $\obs\in\ObsSym$, under the strategy $\osched$.
Let $\pr^*(s)$ denote the probability to satisfy $\ltlformula$ within the induced MC $\pomdp^\osched$.
For some threshold $\lambda'\in[0,1]$, a state $s\in S$ is \emph{critical} iff $\pr^*(s)>\lambda'$. 
We define $\lambda'$ as a function $\lambda'\colon S\times\lambda\rightarrow \R$
with respect to the threshold $\lambda$ from the original specification and the state $s$.
We define the set of critical decision under the strategy $\osched$.

\begin{definition}[Critical Decision]
	A probability $\osched(\obs)(\act)>0$ according to an observation-based strategy $\osched\in \oSched$ is a \emph{critical decision} iff there exist states $s,s'\in S$ with $s\in\ObsFun^{-1}(z)$, $\probmdp(s,\act,s')>0$, and $s'$ is critical.
\end{definition}
Intuitively, a decision is critical if it may lead to a critical state. 
The set of critical decisions serves as \emph{counterexample}, generated by the set of critical states and the strategy $\osched$.
Note that even if a specification is satisfied for $\osched$, the sets of critical decisions and states may still be non-empty as they depend on the definition of the criticality-threshold $\lambda'$.

For each observation $z\in\ObsFun$ with a critical decision, we construct an optimization problem that minimizes the number of different (critical) actions the strategy chooses per observation class.
In particular, the probabilities of action choices under $\osched$ are redistributed such that the critical choices are minimized.
\begin{align}
\label{eq:LP_critic}
&\max_{\osched(\obs)(\act), \act\in \Act} \min_{s\in S} p_s\\
\textit{subject to}&\nonumber\\
\forall s\in\ObsFun^{-1}(z).\ &\quad p_s=
\sum_{\act\in\Act}\osched(\obs)(\act)\cdot\sum_{s'\in S}\probmdp(s,\act,s')\cdot p^*(s')\nonumber
\end{align}
%
%Consider the specification such as \textit{the probability reaching a goal area without bumping into obstacles is above a certain threshold} (see Sect.~\ref{sec:experiments}).
If the objective function is zero, then we have found an observation-based strategy, as there are no choices that are inconsistent with the observations anymore.
Otherwise, we select a class for which at least two different actions are necessary and then we generate a new set of paths starting from the critical states.
After converting these new paths into observation-action sequences, we retrain the \ac{RNN}. By gathering more data from these apparently critical situations, we locally improve the quality of the strategies at those locations and gradually introduce observation-dependencies.
%\nils{synch with 3.2 once done}

%\begin{remark}[Further notions of counterexamples]
%	One may also compute a (minimal) set of states that already suffices to violate the specification, as in~\cite{DBLP:journals/tcs/WimmerJAKB14}. That notion proved less effective in our setting.
%\end{remark}

\subsection{Correctness and Termination}
\emph{Correctness} of our approach is ensured by evaluating the extracted strategy on the \ac{POMDP} using model checking.
As the investigated problem is undecidable for \acp{POMDP}~\cite{MadaniHC99}, our approach is naturally \emph{incomplete}.
In order to enforce termination after finite time, we abort the refinement loop after a specified number of iterations, or as soon as the progress 
from one iteration to the next (in terms of the model checking results) falls below a user-specified threshold.
%\ralf{relative? absolute?}
%\steve{It may be worth taking out that element of the termination condition, since we run a fixed iterative process in the experiments}
% $\varepsilon > 0$, \ie when
%$|p_{i}-p_{i-1}|<\varepsilon$ where $p_i$ is the probability of satisfying the investigated reach-avoid
%probability after iteration $i$.

\section{Experimental Results}
\label{sec:experiments}

We evaluate our \ac{RNN}-based synthesis procedure on benchmark examples that are subject to either \ac{LTL} specifications or expected cost specifications. 
For the former, we compare to the tool \tool{PRISM-POMDP}, and for the latter we compare to \tool{PRISM-POMDP} and the point-based solver \tool{SolvePOMDP}~\cite{DBLP:conf/aaai/WalravenS17}.
We selected the two solvers from different research communities because they provide the possibility for a straightforward adaption to our benchmark setting.
In particular, the tools support undiscounted rewards and have a simple and similar input interface.
Extended experiments with, for instance, Monte-Carlo-based methods~\cite{silver2010monte} are interesting but beyond the scope of this paper.

For a fair comparison, instead of terminating our synthesis procedure once a specification is satisfied, we always iterate 10 times, where one iteration encompasses the (re-)training of the \ac{RNN}, the strategy extraction, the evaluations, and the strategy improvement as detailed in Sect.~\ref{sec:method}.
For instance, for a specification $\reachPropSymbol=\p_{\leq \lambda}(\ltlformula)$, we leave the $\lambda$ open and seek to compute $\p_{\min}(\ltlformula)$, that is, we compute the minimal probability of satisfying $\ltlformula$ to obtain a strategy that satisfies $\reachPropSymbol$.
We cannot guarantee to reach that optimum, but we rather improve as far as possible within the predefined 10 iterations.
The notions are similar for $\p_{\geq \lambda}$ and $\p_{\max}$ as well as for expected cost measures $\Ex_{\leq \lambda}$ ($\Ex_{\geq \lambda}$) and $\Ex_{\min}$ ($\Ex_{\max}$).

We will now shortly describe our experimental setup and present detailed results for both types of examples.
%
%However, in the interest of demonstrating the full capability of the method, the threshold element has been replaced by a fixed number of iterations (10).

%with that seek to maximize satisfying a given LTL specification as well as those that t.

\paragraph{Implementation and Setup.}
%\nj{Put in that subsection: we always train 10 episodes and look where we are then, there is inherent randomness in the \ac{RNN}, but we always use the first shot we get, we could create new strategy without retraining etc.}

%\as{There is no such thing as 'as optimal' or 'less optimal'. It is either optimal or not. Moreover, we do not show that we can handle larger environments than approximate methods. I suggest we rephrase this to: 'The latter example set shows that our proposed method can handle significantly larger environments, with a marginal cost on performance.'}
%The workflow in Fig.~\ref{fig:flowchart} has a stopping criterion of satisfying the specification. 
%However, in the interest of demonstrating the full capability of the method, the threshold element has been replaced by a fixed number of iterations (10).
%The results, when presented, represent the highest quality strategy within 10 improvement update steps.

%We implemented our synthesis on two problem types: a simple \textit{reach-avoid} motion planning setting \cite{carr2018human}; and the well-known problem from POMDP literature \textit{RockSample} \cite{smith2004heuristic}.
%These two problems offer starkly different parameter values for analyzing randomized strategy effectiveness. 
%The motion planning setting has a large observation space $|\ObsSym|=256$  compared to \textit{RockSample} $|\ObsSym|=2$ and demonstrates the limitations of a completely memoryless strategy.
We employ the following \tool{Python} toolchain to realize the full \ac{RNN}-based synthesis procedure.
First, we use the deep learning library \tool{Keras}~\cite{chollet2015keras} to train the strategy network.
To evaluate strategies, we employ the probabilistic model checkers \tool{PRISM} (\ac{LTL}) and \tool{STORM} (undiscounted expected rewards).%\nils{precise: for which specs do you use Storm?}
%Where possible we compare each cumulative reward (or maximum probability of reaching the target) example to 
%Note that there exists no publicly available tool to compute optimal randomized memoryless strategies. 
We evaluated on a 2.3\,GHz machine with a 12\,GB memory limit and a specified maximum computation time of $10^5$~seconds.
%\alex{It would be nice to also mention an average number of training steps. E.g. for the small environments it took 15 training steps, for the large ones 200, etc.}

\subsection{Temporal Logic Examples}
%\paragraph{Setting Environment.}
We examined three problem settings involving motion planning with \ac{LTL} specifications.
For each of the settings, we use a standard gridworld formulation of an agent with 4~action choices (cardinal directions of movement), see Fig.~\ref{fig:ExampleEnvironment}.
Inside this environment there are a set of static $(\hat{x})$ and moving $(\tilde{x})$ obstacles as well as possible target cells $A$ and $B$. %,C \rbrace$.
Each agent has a limited visibility region, indicated by the green area, and can infer its state from observations and knowledge of the environment.
% (static obstacles and target states).
We define observations as Boolean functions that take as input the positions of the agent and moving obstacles.
Intuitively, the functions describe the 8 possible relative positions of the obstacles with respect to the agent inside its viewing range. 
%
%
%the observation model is given by a set of $8$ Boolean classifying functions as a vector. 
%If the agent is adjacent to a wall or an obstacle, then the equivalent Boolean function returns \true for that index in the observation vector.
%The observation space $\ObsSym = \lbrace z_0,\ldots,z_{255} \rbrace$ is 8-bit set of all observation vectors.
%%\nj{put sentence about why we minimize/maximize here.}
%\spc{Mentioned above in the reason why we iterate 10 times. Should I repeat or just reference or neither?}
%
%
\begin{figure}
\setlength{\tabcolsep}{2pt}
\centering
\subfigure[\label{fig:ExampleEnvironment}]{\newcommand{\MovingObstacle}[2]{
% \fill[red] (#1+0.15,#2+0.15) rectangle (#1+0.85,#2+0.85);
\fillGridAt{#1}{#2}{\Huge \color{red} $\tilde{x}$}
 \draw[thick,red,->,shorten >=2pt,shorten <=2pt,>=stealth] (#1+0.5,#2+0.85) -- (#1+0.5,#2+1.35);
\draw[thick,red,->,shorten >=2pt,shorten <=2pt,>=stealth] (#1+0.75,#2+0.5) -- (#1+1.25,#2+0.5);
\draw[thick,red,->,shorten >=2pt,shorten <=2pt,>=stealth] (#1+0.25,#2+0.5) -- (#1-0.25,#2+0.5);
\draw[thick,red,->,shorten >=2pt,shorten <=2pt,>=stealth] (#1+0.5,#2+0.25) -- (#1+0.5,#2-0.25);
}

\newcommand{\MovingObject}[2]{
	\fill[blue] (#1+0.15,#2+0.15) rectangle (#1+0.85,#2+0.85);
	\draw[thick,blue,->,shorten >=2pt,shorten <=2pt,>=stealth] (#1+0.5,#2+0.75) -- (#1+0.5,#2+1.25);
	\draw[thick,blue,->,shorten >=2pt,shorten <=2pt,>=stealth] (#1+0.75,#2+0.5) -- (#1+1.25,#2+0.5);
	\draw[thick,blue,->,shorten >=2pt,shorten <=2pt,>=stealth] (#1+0.25,#2+0.5) -- (#1-0.25,#2+0.5);
	\draw[thick,blue,->,shorten >=2pt,shorten <=2pt,>=stealth] (#1+0.5,#2+0.25) -- (#1+0.5,#2-0.25);
	\fill [green,opacity=0.25]  (#1-1,#2-1) rectangle (#1+2,#2+2);
}

\newcommand{\StaticObstacle}[2]{\fillGridAt{#1}{#2}{\Huge \color{red} $\hat{x}$}}

\scalebox{0.55}{
\begin{tikzpicture}
\draw[black,line width=1pt] (0,0) grid[step=1] (5,5);
\draw[black,line width=4pt] (0,0) rectangle (5,5);
\fillGridAt{4}{4}{\Huge $A$}
\fillGridAt{0}{0}{\Huge $B$}
%\fillGridAt{0}{4}{\Huge $C$}

\StaticObstacle{1}{0}
\StaticObstacle{1}{2}

%\StaticObstacle{2}{1}
\MovingObstacle{3}{1}
\MovingObject{2}{3}
\end{tikzpicture}
}}
\hfill
\subfigure[\label{tab:probmetrics}]{\raisebox{1.8cm}{%
    \scriptsize%
    \begin{tabular}{@{}lrrr@{}}
	\toprule
	Problem & $|S|$ & $|\Act|$ & $|\ObsSym|$\\
	\midrule
	\rowcolor{gray!25} Navigation ($\gridsizeparam$) & $\gridsizeparam^4$ & 4 &\ 256 \\
	\rowcolor{gray!25} Delivery ($\gridsizeparam$) & $\gridsizeparam^2$ & 4 & 256 \\
	\rowcolor{gray!25} Slippery ($\gridsizeparam$) & $\gridsizeparam^2$ & 4 & 256 \\
	Maze($\gridsizeparam$) & $3\gridsizeparam +8 $  & 4 & 7\\
	Grid($\gridsizeparam$) & $\gridsizeparam^2$ & 4 & 2 \\
%	Slippery &  11 & 4 & 6 \\
	RockSample$[4,4]$ & 257 & 9  & 2\\
	RockSample$[5,5]$ & 801 & 10 & 2\\
	RockSample$[7,8]$ & 12545 & 13 & 2\\
	\bottomrule
    \end{tabular}}}
    \caption{(a) Example environment and (b) Benchmark metrics}
\end{figure}
%
%\paragraph{Setting Specifications.}
%\label{ssec:Specs}
%We examined three different motion planning scenarios, each with slightly different environments and specifications.
%
%
%seek $\max(\lambda)$ s.\,t. $\lambda \in [0,1]$ for $\varphi_1$ and $\varphi_3$ as well as $\min(\lambda)$ s.\,t. $\lambda \in \left[ 0,\infty \right)$ for $\varphi_2$:
\begin{enumerate}
	\item \textbf{Navigation with moving obstacles} -- an agent and a single stochastically moving obstacle. 
	The agent task is to maximize the probability to navigate to a goal state $A$ while not colliding with obstacles (both static and moving): 
	 $\varphi_1= \p_{\max}\left( \neg X\,\Until\,A\right)$ with $x = \hat{x}\cup\tilde{x}$,
	\item \textbf{Delivery without obstacles} -- an agent and static objects (landmarks). The task is to deliver an object from $A$ to $B$ in as few steps as possible: $\varphi_2= \Ex_{\min}(\finally(A \land \Ever B))$.
	\item \textbf{Slippery delivery with static obstacles} -- an agent where the probability of moving perpendicular to the desired direction is $0.1$ in each orientation. 
	The task is to maximize the probability to go back and forth from locations $A$ and $B$ without colliding with the static obstacles $\hat{x}$: $ \varphi_3= \p_{\max} \left( \Always \Ever A \land \Always \Ever B \land \neg \Ever X \right)$, with $x = \hat{x}$,
%	\item \textbf{Surveillance} -- a single agent with dynamics similar to \emph{slippery delivery} and a static set of obstacles. In this setting, the agent must patrol between $C$ and $A$ or $C$ and $B$ above a prescribed probability $\lambda$.
%	Formally denoted by:  $ \varphi_4= \pr_{\max}^\pomdp \left(\Always \Ever C \land (\Always \Ever A \lor \Always \Ever B)\land \neg \Ever X\right) $, where $X = \hat{x}$.
\end{enumerate}

%In computing the strategy, the procedure is concerned with finding the strategy that maximizes (or minimizes in the case of expected cost specification $\varphi_2$) the probability that the specification is satisfied.
%In these cases the stopping criterion of SAT in Fig.~\ref{fig:flowchart} is not sufficient and so the workflow is run for 10 iterations before stopping.
%This was chosen to fully demonstrate the capability of the procedure to these more complicated settings.
%Some of settings ($\varphi_1$) have specifications that could be converted into PCTL with a probability threshold that cuts off when it is satisfied, resulting in the behavior described in the workflow in Fig.~\ref{fig:flowchart}.

\begin{figure}[ht]
	\centering
	\scalebox{0.79}{\begin{tikzpicture}
	\begin{axis}[
	axis y line*=left,
	title = {Refinement statistics},
	xlabel = {Iteration no.},
	ylabel = {\# critical states},
	ymin= 0,
	ymax=1500,
	xmin=0,
	xmax=10,
		label style={font=\small},
	tick label style={font=\small},
	width = 0.4\textwidth,
	height = 0.32\textwidth]
	\addplot+[mark=x,only marks,mark size = 3.5pt]coordinates{(1,1221)(2,907)(3,811)(4,751)(5,706)(6,643)(7,604)(8,589)(9,561)};\label{plot_count}
	%\addlegendentry{Counterexamples}
	\end{axis};
	\begin{axis}[
	axis y line=right,
	axis x line=none,
	ylabel = {$\pr(\neg X \Until A)$},
	ymin= 0,
	ymax=1,
	xmin=0,
	xmax=10,
	width = 0.4\textwidth,
	height = 0.32\textwidth,
	label style={font=\small},
	tick label style={font=\small},
	legend style={at={(1,0)},anchor=south east,font=\small}]
	\addplot+[red,mark=*,mark options={red},only marks,mark size = 3.5pt]coordinates{(1,0.156)(2,0.509)(3,0.680)(4,0.775)(5,0.845)(6,0.896)(7,0.922)(8,0.937)(9,0.948)};\label{plot_reach}
	\addlegendimage{/pgfplots/refstyle=plot_count,}
	\addlegendentry{Probability}
	\addlegendentry{Counterexamples}
	\end{axis};
	
	\end{tikzpicture}}
	\caption{Progression of the number of critical states and the probability of satisfying an LTL specification as a result of local improvement steps.
%	reaching the target $A$ without crashing. This is a single instance of \emph{Navigation}(6)
}
	\label{fig:Refinement}
\end{figure}
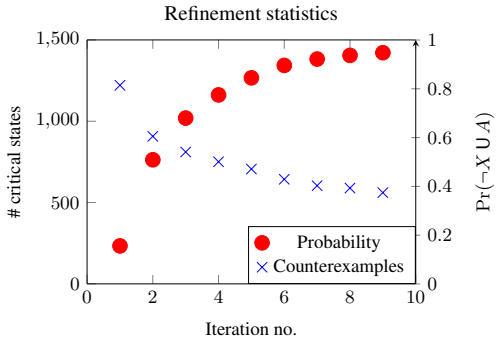
%\alex{This graph reads bit weird: if we satisfy the property with probability 1, why do we have so many counter examples?}

%The metrics associated with training the RNN are given in Table~\ref{tab:Parameters}.
%The number of samples generally scales with the number of states, however, for environments larger than the $10 \times 10$ grid we utilize the samples from smaller environments to train significantly larger state spaces.
%The predictive accuracy of the network to map the sample data to an action choice is fairly consistent across all problem sizes.
%
%\begin{table}[t]
%		\centering
%		\caption{Navigation problem -- training metrics}
%		\label{tab:Parameters}
%		\centering
%		\scalebox{0.9}{
%			\begin{tabular}{lcr}
%				\toprule
%				Grid-size & No. of samples & Categorical cross-entropy accuracy \\
%				\midrule
%				4 & 1534 & 0.844 \\
%				5 & 3231 & 0.894 \\
%				6 & 7123 & 0.889 \\
%				8 & 20580 & 0.914 \\
%				10 & 434210 & 0.852 \\
%				15 & 21340* & 0.893 \\
%				20 & 22023* & 0.895 \\
%				30 & 20394* & 0.896 \\
%				
%				\bottomrule
%		\end{tabular}}
%\end{table}

\paragraph{Evaluation.}
Fig.~\ref{fig:Refinement} compares the size of counterexample in relation to the probability of satisfying an LTL formula in each iteration of the synthesis procedure. 
In particular, we depict the size of the set $S'\subset S$ of critical states regarding $\varphi_1=\p_{\max}\left( \neg X\,\Until\,A\right)$ for the $\mathit{Navigation}$ example with grid-size 6.
Note that even if the probability to satisfy the LTL specification is nearly one (for the initial state of the POMDP), there may still be critical intermediate states.
As can be seen in the figure, while the probability to satisfy the LTL formula increases, the size of the counterexample decreases. 
In particular, the local improvement (Eq.~\ref{eq:LP_critic}, Sect.~\ref{ssec:drnn_improve}) is demonstrated to be effective.
%
%
%that are classified as ``critical" for $\varphi_1$ in a $\mathit{Navigation}(6)$ example for 10 iterations.
%The 
%As the procedure produces additional counterexamples, the local improvement (see Eq.~\ref{eq:LP_critic}) and additional training data steps help improve the extracted strategy's probability of satisfying the reach-avoid specification.

Table~\ref{tab:LTL_Prop} contains the results for the above LTL examples.
Naturally the strategies produced by the procedure will not have higher maximum probabilities (or lower minimum expected cost) than those generated by the PRISM-POMDP tool.
However, they scale for significantly larger environments and settings.
In the larger environments ($\mathit{Navigation}(15)$ and upwards indicated by a star) we employ the sampling technique outlined at the end of Sect.~\ref{ssec:DRNN} on a dataset with grid-size 10.
The strategy still scales to these larger environments even when trained on data from a smaller state space.

%\paragraph{Comparison of Multiple Strategy Types.}
Also in Table~\ref{tab:LTL_Prop}, we compare the effect of increasing the value of $k$ for several $k$-\acp{FSC}.
In smaller instances with grid-sizes of 4 and 5, memory-based strategies significantly outperform memoryless ones in terms of quality (the resulting probability or expected cost) while not consuming significantly more time.
The increase in performance is due to additional expressiveness of an FSC-based strategy in these environments with a higher density of obstacles.

Summarized, our method scales to significantly larger domains than \tool{PRISM-POMDP} with competitive computation times.
%Especially for such large examples, the computation times are competitive.
As mentioned before, there is an inherent level of randomness in extracting a strategy.
While we always take the first shot result for our experiments, the quality of strategies may improved by sampling several \ac{RNN} predictions.

%For further improvement, better sampling could be conducted with some refinements to the counterexamples procedure, however, the gains would be minimal compared the data required to achieve the performance.

%Even in the larger gridworlds, 
%In larger gridworlds, where the sampled data more readily covers the entire state-memory product space, the larger values of $k$ in $k$-\ac{FSC} produce much higher quality strategies than the simpler memory sequence counterparts.
%\spc{Replace this when we get a better sense of the result.}

\begin{table}[tb]
	\scriptsize\setlength{\tabcolsep}{2pt}
	\caption{Synthesizing strategies for examples with LTL specs.}
	\label{tab:LTL_Prop}
	\centering
	\scalebox{0.95}{\begin{tabular}{@{}lrc|rr|rr@{}}
		\toprule
		&&& \multicolumn{2}{c}{RNN-based Synthesis} &  \multicolumn{2}{c}{PRISM-\ac{POMDP}}  \\
		Problem & States     & Type, $\varphi$    &  Res. & Time (s)  &  Res. & Time (s)  \\
		\midrule
		Navigation (3)& 333 &$\p_{\max}^\pomdp$, $\varphi_1$  & 0.74 & \textbf{14.16}  & \textbf{0.84} & 73.88 \\
		\rowcolor{gray!25} Navigation (4) & 1088 &$\p_{\max}^\pomdp$, $\varphi_1$ & 0.82 & \textbf{22.67}  & \textbf{0.93} & 1034.64 \\
		\rowcolor{gray!25} Navigation (4) [2-FSC] & 13373  & $\p_{\max}^\pomdp$, $\varphi_1$  & 0.91 & 47.26 & -- & -- \\
		\rowcolor{gray!25} Navigation (4) [4-FSC] & 26741 & $\p_{\max}^\pomdp$, $\varphi_1$  & 0.92 & 59.42 & -- & -- \\
		\rowcolor{gray!25} Navigation (4) [8-FSC]  & 53477  & $\p_{\max}^\pomdp$, $\varphi_1$ & \textbf{0.92} & 85.26  & -- & -- \\
		Navigation (5) & 2725 &$\p_{\max}^\pomdp$, $\varphi_1$  & 0.91 & \textbf{34.34}  & MO & MO  \\
		Navigation (5) [2-FSC] &  33357 & $\p_{\max}^\pomdp$, $\varphi_1$  & 0.92 & 115.16  & -- & -- \\
		Navigation (5) [4-FSC] & 66709  & $\p_{\max}^\pomdp$, $\varphi_1$  & 0.92 & 159.61  & -- & -- \\
		Navigation (5) [8-FSC] & 133413 & $\p_{\max}^\pomdp$, $\varphi_1$   & \textbf{0.92} & 250.91  & -- & -- \\
%		Navigation (6)  & 6020 &$\mathbb{P}_{\max}$, $\varphi_1$ & 0.96 & 75.05 & 4896 & MO & MO \\
		\rowcolor{gray!25} Navigation (10) & 49060 &$\p_{\max}^\pomdp$, $\varphi_1$ & 0.79 & \textbf{822.87}  & MO & MO \\
		\rowcolor{gray!25} Navigation (10) [2-FSC] &  475053  & $\p_{\max}^\pomdp$, $\varphi_1$   & 0.83 & 1185.41  & -- & -- \\
		\rowcolor{gray!25} Navigation (10)  [4-FSC] &  950101 & $\p_{\max}^\pomdp$, $\varphi_1$   & \textbf{0.85} & 1488.77 & -- & -- \\
		\rowcolor{gray!25} Navigation (10) [8-FSC]  & 1900197 & $\p_{\max}^\pomdp$, $\varphi_1$    & 0.81 & 1805.22 & -- & -- \\
		Navigation (15) & 251965 &$\p_{\max}^\pomdp$, $\varphi_1$  & \textbf{0.91} & \textbf{1271.80}*  & MO & MO \\
		Navigation (20) & 798040 &$\p_{\max}^\pomdp$, $\varphi_1$  & \textbf{0.96}& \textbf{4712.25}*  & MO & MO \\
		Navigation (30) & 4045840 &$\p_{\max}^\pomdp$, $\varphi_1$  & \textbf{0.95} & \textbf{25191.05}*  & MO & MO \\
		Navigation (40) & -- &$\p_{\max}^\pomdp$, $\varphi_1$ & TO & TO & MO & MO \\
		\hdashline
		Delivery (4) [2-FSC] & 80 &$\Ex_{\min}^\pomdp$, $\varphi_2$  &	 6.02 	&35.35   & \textbf{6.0} & \textbf{28.53} \\
		Delivery (5) [2-FSC] & 125  &$\Ex_{\min}^\pomdp$, $\varphi_2$  & 8.11 & \textbf{78.32}  & \textbf{8.0} & 102.41 \\
		Delivery (10) [2-FSC] & 500 &$\Ex_{\min}^\pomdp$, $\varphi_2$  & \textbf{18.13} & \textbf{120.34} & MO & MO \\
		\hdashline
		Slippery (4) [2-FSC]& 460 & $\p_{\max}^\pomdp$, $\varphi_3$  & 0.78  & 67.51  & \textbf{0.90} & \textbf{5.10} \\
		Slippery (5) [2-FSC]& 730 & $\p_{\max}^\pomdp$, $\varphi_3$  & 0.89 & 84.32   & \textbf{0.93} & \textbf{83.24}  \\
		Slippery (10) [2-FSC]& 2980 & $\p_{\max}^\pomdp$, $\varphi_3$  & \textbf{0.98} & \textbf{119.14}    & MO & MO \\
		Slippery (20) [2-FSC] & 11980 & $\p_{\max}^\pomdp$, $\varphi_3$  & \textbf{0.99} & \textbf{1580.42}    & MO & MO \\
%		\hdashline
%		Surveillance (4) & $\mathbb{P}_{\max}$ , $\varphi_4$  & 384 & 0.00 & 39.78 & 107 & 0.90 & 4.91  \\
%		Surveillance (5) & $\mathbb{P}_{\max}$ , $\varphi_4$  &714 & 0.80 & 84.78 & 149 & 0.90 &   \\
		\bottomrule
	\end{tabular}}
\end{table}

\subsection{Comparison to Existing \ac{POMDP} Examples}
%\spc{TODO:Simple description of RockSample, Grid and Slippery}

For comparison to existing benchmarks, we extend two examples from PRISM-\ac{POMDP} for an arbitrary-sized structure: \emph{Maze}($\gridsizeparam$) with $\gridsizeparam+2$ rows  and \emph{Grid}($\gridsizeparam$) -- a square grid with length $\gridsizeparam$. 
We also compare to \emph{RockSample}~\cite{silver2010monte} (see Table~\ref{tab:probmetrics} for problem metrics). 

These problems are quite different to the LTL examples, in particular the significantly smaller observation spaces.
As a result, a simple memoryless strategy is insufficient for a useful comparison.
For each problem, the size of the $k$-FSC used is given by: Maze(\gridsizeparam) has $k=(\gridsizeparam+1)$; Grid(\gridsizeparam) has $k= (\gridsizeparam-1)$ and \emph{RockSample} with $b$ rocks has $k=b$.

Our method compares favorably with \tool{PRISM-POMDP} and \tool{pomdpSolve} for Maze and Grid (Table~\ref{tab:ToolComp}).
However, the proposed method performs poorly in comparison to \tool{pomdpSolve} for \emph{RockSample}:
An observation is received after taking an action to \emph{check} a particular rock. 
This action is never sampled in the modified trajectory-tree based sampling method (Sect.~\ref{ssec:DRNN}).
%To create a dataset we utilize an adjustable parameter for the probability of checking a rock. 
Note that our main aim is to enable the efficient synthesis of strategies under linear temporal logic constraints.
%An alternate approach may involve constructing the dataset from expert demonstrations, however, such a set would need to be large and as such may take a significantly long time to synthesize.
%Please note, these benchmarks are not the purpose of the proposed procedure, they are merely useful as a comparative tool for the existing solvers that are not capable of synthesizing strategies for formal specifications.

\begin{table}[tb]
	\scriptsize\setlength{\tabcolsep}{2pt}
	\caption{Comparison for standard POMDP examples.}
	\label{tab:ToolComp}
	\centering
	\scalebox{0.95}{%
		\begin{tabular}{@{}cc|rrr|rr|rr@{}}
			\toprule
			& & \multicolumn{3}{c}{RNN-based Synthesis}   &\multicolumn{2}{c}{PRISM-\ac{POMDP}} & \multicolumn{2}{c}{\tool{pomdpSolve}}  \\
			Problem      & Type   & States & Res & Time (s)& Res & Time (s) & Res & Time (s) \\
			\midrule
			Maze (1) &$\Ex_{\min}^\pomdp$  & 68 &   4.31 &   31.70 & \textbf{4.30} & \textbf{0.09} & 4.30 & 0.30  \\
			Maze (2) &$\Ex_{\min}^\pomdp$   & 83 &	5.31 	& 46.65 &   5.23  & 2.176 & \textbf{5.23} & \textbf{0.67}  \\
			Maze (3) &$\Ex_{\min}^\pomdp$ & 98  & 	8.10 	&  58.75 	 	& 7.13    & 38.82 &  \textbf{7.13} & \textbf{2.39} \\
			Maze (4) &$\Ex_{\min}^\pomdp$ & 113  &   11.53    & 58.09     & 8.58 & 543.06 & \textbf{8.58} & \textbf{7.15} \\
			Maze (5) &$\Ex_{\min}^\pomdp$ & 128  &     14.40    &   \textbf{68.09}    & 13.00 & 4110.50 & \textbf{12.04} & 132.12 \\
			Maze (6) &$\Ex_{\min}^\pomdp$ &  143 &    22.34   & \textbf{71.89}     & MO & MO & \textbf{18.52} & 1546.02 \\
			Maze (10)&$\Ex_{\min}^\pomdp$ & 203 &  100.21 & \textbf{158.33} & MO & MO &  MO & MO \\
			\hdashline
			Grid (3) &$\Ex_{\min}^\pomdp$ & 165 &  2.90 &  38.94 & 2.88  & 2.332 & \textbf{2.88} & \textbf{0.07} \\
			Grid (4)  &$\Ex_{\min}^\pomdp$ & 381   &  4.32	 	&  	79.99 	& 4.13    &  1032.53  & \textbf{4.13} & \textbf{0.77} \\
			Grid (5) &$\Ex_{\min}^\pomdp$ & 727 & 6.623 & 91.42 & MO & MO & \textbf{ 5.42}  & \textbf{1.94}\\
			Grid (10) &$\Ex_{\min}^\pomdp$ & 5457  & \textbf{13.630} & \textbf{268.40} & MO & MO & MO & MO \\
%			\hdashline
%			Slippery (4) &$\p_{\max}^\pomdp$ &  56   & 	0.90	&	120.32 	& MO & MO  &0.93    & 7637.42   \\
			\hdashline
			RockSample$[4,4]$ & $\Ex_{\max}^\pomdp$ &  2432  & 	17.71 	&	35.35 	& N/A & N/A  & \textbf{18.04}  & \textbf{0.43}  \\
			RockSample$[5,5]$ &$\Ex_{\max}^\pomdp$  &  8320  &	18.40	& 	\textbf{43.74} & N/A & N/A 	&  \textbf{19.23}   &  621.28  \\
			RockSample$[7,8]$ & $\Ex_{\max}^\pomdp$ &  166656   &	 20.32	& \textbf{860.53} & N/A & N/A 	& \textbf{21.64}    & 20458.41   \\
			\bottomrule
	\end{tabular}}
\end{table}

\section{Summary and Future Work}
We introduced a new \ac{RNN}-based strategy synthesis method for \acp{POMDP} and LTL specifications.
While we cannot guarantee optimality, our approach shows results that are often close to the actual optimum with competitive computation times for large problem domains.

For the future, we are interested in extending our method to continuous state spaces together with abstraction techniques that would enable to employ our model-based method.

%\clearpage
\bibliographystyle{named}
\bibliography{literature}

\end{document}